\pdfoutput=1

\documentclass[11pt]{article}

\usepackage{acl}

\usepackage{times}
\usepackage{latexsym}

\usepackage[T1]{fontenc}

\usepackage[utf8]{inputenc}

\usepackage{microtype}

\usepackage{inconsolata}

\usepackage{graphicx}

\usepackage{accents}
\newlength{\dhatheight}

\newcommand{\doublehat}[1]{%
    \settoheight{\dhatheight}{\ensuremath{\hat{#1}}}%
    \addtolength{\dhatheight}{-0.25ex}%
    \hat{\vphantom{\rule{1pt}{\dhatheight}}%
    \smash{\hat{#1}}}}

\newcommand{\rparagraph}[1]{\vspace{1.4mm}\noindent\textbf{#1.}}
\newcommand{\rrparagraph}[1]{\vspace{1.2mm}\noindent\textit{#1.}}
\newcommand{\sparagraph}[1]{\vspace{0.0mm}\noindent\textbf{#1.}}

\newcommand{\ttr}{{\texttt{T-Train}}\xspace}
\newcommand{\tts}{{\texttt{T-Test}}\xspace}
\newcommand{\ett}{{\texttt{ETT}}\xspace}

\newcommand{\zs}{ZS\xspace}
\newcommand{\xlt}{XLT\xspace}

\newcommand{\mlm}{mLM\xspace}
\newcommand{\mlms}{mLMs\xspace}
\newcommand{\wa}{WA\xspace}
\newcommand{\was}{WAs\xspace}
\newcommand{\ep}{\texttt{Easy}\xspace}
\newcommand{\codec}{\texttt{Codec}\xspace}
\newcommand{\ept}{Easy\xspace}
\newcommand{\codect}{Codec\xspace}

\newcommand{\llmvec}{LLM2Vec\xspace}
\newcommand{\comsrc}{\texttt{COMP-SRC}}
\newcommand{\comtgt}{\texttt{COMP-TGT}}
\newcommand{\comins}{\texttt{COMP-INS}}
\newcommand{\corsch}{\texttt{RSTR-TGT}}
\newcommand{\nofilt}{\texttt{NO-FILT}}
\newcommand{\wstok}{\texttt{WS-TOK}\xspace}
\newcommand{\sptok}{\texttt{LS-TOK}\xspace}

\newcommand{\tcomsrc}{COMP-SRC\xspace}
\newcommand{\tcomtgt}{COMP-TGT\xspace}
\newcommand{\tcomins}{COMP-INS\xspace}
\newcommand{\tcorsch}{RSTR-TGT\xspace}
\newcommand{\tnofilt}{NO-FILT\xspace}
\newcommand{\twstok}{WS-TOK\xspace}
\newcommand{\tsptok}{LS-TOK\xspace}

\usepackage{multirow}
\usepackage{adjustbox}
\usepackage{booktabs}
\usepackage{tabularx}
\usepackage{verbatim}
\usepackage{hyperref}
\usepackage{xspace}
\usepackage{caption}
\usepackage{subcaption}
\usepackage{amssymb}

\title{The Devil Is in the Word Alignment Details: On Translation-Based Cross-Lingual Transfer for Token Classification Tasks}

\author{Benedikt Ebing and Goran Glava\v{s} \\
  University of W\"{u}rzburg \\ Center for Artificial Intelligence and Data Science (CAIDAS) \\
  \texttt{\{benedikt.ebing, goran.glavas\}@uni-wuerzburg.de} \\}

\begin{document}
\maketitle

\begin{abstract}
Translation-based strategies for cross-lingual transfer (\xlt) such as \textit{translate-train}---training on noisy target language data translated from the source language---and \textit{translate-test}---evaluating on noisy source language data translated from the target language---are competitive \xlt baselines. 
In \xlt for token classification tasks, however, these strategies include 
\textit{label projection}, the challenging step of mapping the labels from each token in the original sentence to its counterpart(s) in the translation.
Although word aligners (\was) are commonly used for label projection, the low-level design decisions for applying them to translation-based \xlt have not been systematically investigated. Moreover, recent marker-based methods, which project labeled spans by inserting tags around them before (or after) translation, claim to outperform \was in label projection for \xlt.
In this work, we revisit \was for label projection, systematically investigating the effects of low-level design decisions on token-level \xlt: (i) the algorithm for projecting labels between (multi-)token spans, 
(ii) filtering strategies to reduce the number of noisily mapped labels, and (iii) the pre-tokenization of the translated sentences. We find that all of these substantially impact translation-based \xlt performance and show that, with optimized choices, \xlt with \wa offers performance at least comparable to that of marker-based methods.  
We then introduce a new projection strategy that ensembles \textit{translate-train} and \textit{translate-test} predictions and demonstrate that it substantially outperforms the marker-based projection. Crucially, we show that our proposed ensembling also reduces sensitivity to low-level \wa design choices, resulting in more robust \xlt for token classification tasks.
\end{abstract}

\section{Introduction}
In recent years, multilingual language models (\mlms) have \textit{de facto} become the main vehicle of cross-lingual transfer (\xlt): fine-tuned on labeled task data in a high-resource source language, \mlms can make predictions in target languages with few (few-shot \xlt) to no (zero-shot \xlt) labeled task instances \cite{wu-dredze-2019-beto, wang2019cross,lauscher-etal-2020-zero, schmidt-etal-2022-dont}. 
While both encoder-only \cite{devlin-etal-2019-bert, conneau-etal-2020-unsupervised, he2023debertav3improvingdebertausing} and decoder-only \cite{gemmateam2024gemma2improvingopen, hui2024qwen25codertechnicalreport, grattafiori2024llama3herdmodels} \mlms have demonstrated strong \xlt performance for sequence classification tasks, in \xlt for token classification tasks the comparatively smaller encoder-only \mlms, like XLM-R \cite{conneau-etal-2020-unsupervised}, continue to outperform the much larger decoder \mlms \cite{ahuja-etal-2023-mega, le2024constraineddecodingcrosslinguallabel, parekh-etal-2024-contextual}.
Much of the above work highlights translation-based strategies as competitive approaches for \xlt, where a machine translation (MT) model is used to either (1) generate noisy target language data by translating the original source language data before training, known as \textit{translate-train} (\ttr), or (2) translate original target language instances into the (noisy) source language before inference, known as \textit{translate-test} (\tts) \cite{pmlr-v119-hu20b, ruder-etal-2021-xtreme, ebrahimi-etal-2022-americasnli, aggarwal-etal-2022-indicxnli}. More elaborate translation-based \xlt strategies have recently been shown to further improve the transfer performance \cite{artetxe-etal-2023-revisiting,ebing-glavas-2024-translate}.   



The effectiveness of translation-based \xlt, however, has predominantly been showcased on sequence-level classification tasks \cite{ruder-etal-2021-xtreme, oh-etal-2022-synergy, artetxe-etal-2023-revisiting}. This is in part due to the fact that translation-based \xlt for \textit{token classification tasks} entails the (difficult) step of \textit{label projection}. Traditionally, label projection is tackled with word aligners (\was) \cite{och-ney-2003-systematic, dyer-etal-2013-simple, dou-neubig-2021-word}, which map each token in the source sequence to a corresponding token in the target sequence. Recent \wa work leverages contextualized embeddings from \mlms (e.g., mBERT) to produce token alignments \cite{jalili-sabet-etal-2020-simalign, dou-neubig-2021-word, wang-etal-2022-multilingual}.
Although \wa research has a long-standing track record in NLP \cite{och-ney-2003-systematic, dyer-etal-2013-simple, jalili-sabet-etal-2020-simalign, dou-neubig-2021-word, wang-etal-2022-multilingual}, standard \wa evaluation protocols do not include translation-based \xlt for token classification tasks. 
Because of this, the impact of low-level design decisions related to token-level \xlt using \was---such as (i) the exact algorithm for projecting the labels, (ii) filtering techniques to reduce the number of noisily mapped labels, and (iii)
the process of inducing word boundaries (i.e. pre-tokenization) to the translated sentence before it can be aligned to the tokens in the original sentence---remain underinvestigated \cite{dou-neubig-2021-word, garcia-ferrero-etal-2022-model}.

In the meantime, marker-based label projection \cite{chen-etal-2023-frustratingly, le2024constraineddecodingcrosslinguallabel} has largely replaced \wa for label projection for token-level \xlt. 
These approaches insert tags (e.g., "\textbf{[}", "\textbf{]}") around labeled spans of interest, either (i) before translation to preserve the markers throughout the translation process and recover the spans afterward \cite{chen-etal-2023-frustratingly}, or (ii) post-translation by means of constrained decoding \cite{le2024constraineddecodingcrosslinguallabel}.\footnote{Further methods have been proposed that address label projection by constrained or contextualized generation of labeled spans given the translated input sentence \cite{garcia-ferrero-etal-2023-projection, parekh-etal-2024-contextual}. We provide details on these methods in App.~\ref{app:tproj} and \ref{app:clap}. In our preliminary experiments these methods performed at most on par with Codec (see App.~\ref{app:ablation_proj}).} 

This line of work explicitly evaluates token-level translation-based \xlt, demonstrating strong performance for both \ttr and \tts. Furthermore, it renders \wa-based \xlt for token classification tasks inferior \cite{chen-etal-2023-frustratingly, le2024constraineddecodingcrosslinguallabel}.   
While these efforts provide the technical details for their proposed marker-based methods, they do not lay out the low-level design details for label projection with \was. Therefore, we argue that they possibly underestimate translation-based \xlt with \was due to suboptimal design choices.

\rparagraph{Contributions} Because of this, we \textbf{(1)} systematically investigate \wa for token-level translation-based \xlt. We start by evaluating the effect of low-level design decisions covering (i) the exact algorithm for mapping the labels between \mbox{(multi-)token} spans from the source sentence to the target sentence based on word alignments; (ii) filtering strategies to identify incomplete labeled span alignments, and 
(iii) the process of inducing word boundaries (i.e. pre-tokenization) to the translated sentence required to align the tokens to their counterparts in the original sentence.
We demonstrate that these design choices can substantially impact translation-based \xlt performance for token-level tasks. For example, using a language-specific pre-tokenizer instead of simple whitespace pre-tokenization improves the performance of \tts by up to 12.6\%.
Overall, we find \tts to be more sensitive to low-level design decisions of \wa than \ttr. 
\textbf{(2)} We then extensively compare \wa-based label projection---with the identified well-performing low-level design choices---against state-of-the-art marker-based label projection methods in token-level \xlt 
on two established benchmarks encompassing 29 
diverse languages. 
Contrary to prior claims \cite{chen-etal-2023-frustratingly, le2024constraineddecodingcrosslinguallabel,parekh-etal-2024-contextual}, we find that optimized \wa-based label projection matches or surpasses the performance of marker-based approaches in translation-based \xlt on token-level tasks. 
\textbf{(3)} Moreover, we propose a more sophisticated method for token-level translation-based \xlt with \was based on ensembling \ttr and \tts \cite{oh-etal-2022-synergy}. For each token, we average the probability distributions over the labels produced by \ttr and \tts. Our ensemble (\ett) improves the transfer performance substantially, outperforming state-of-the-art marker-based approaches. More importantly, \ett drastically reduces the sensitivity of \ttr and \tts to low-level \wa design decisions. \textbf{(4)} Finally, we show that our findings hold for different MT models (i.e., translation quality), \wa models, and base LMs (i.e., encoder vs. decoder models). We publicly release our code and data: \href{https://github.com/bebing93/devil-in-details}{https://github.com/bebing93/devil-in-details}.


\section{Token-Level XLT via Word Alignment}
\label{sec:preliminary}
\begin{figure*}[t!]
\centering
\includegraphics{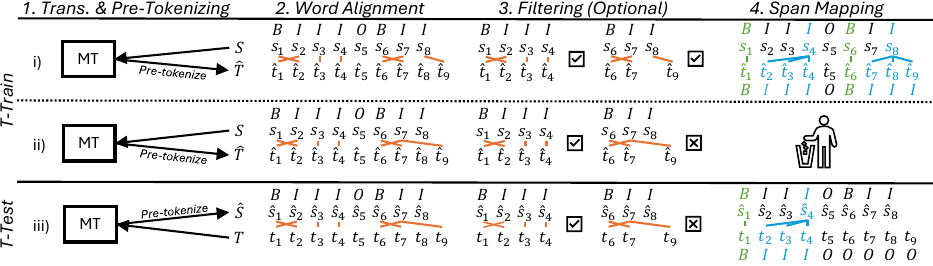} 
\caption{Schematic overview of our pipeline for label projection using word aligners. From top to bottom, we display three cases---two for \ttr and one for \tts: (i) a successful creation of a translated target language training instance $\hat{T}$ from the original source language training instance $S$, (ii) a discarded translated target language training instance $\hat{T}$ due to incomplete alignments detected by our filtering strategies, and (iii) a partial label projection from a translated source language test instance $\hat{S}$ to the original target language test instance $T$ due to incomplete alignments detected by our filtering strategies. All cases exemplarily apply the \textit{Complete Source} filter. 
}
\label{fig:entityMatch}
\vspace{-0.5em}
\end{figure*}
We first detail our pipeline for label projection with \wa focusing on the low-level design choices we investigate: mapping labeled spans, filtering strategies, and pre-tokenization \cite{garcia-ferrero-etal-2022-model}. We then describe our new translation-based XLT approach for token-level tasks that ensembles \ttr and \tts.

\subsection{Label Projection with \wa: Design Choices}
\label{ssec:design_choices}
Translation-based XLT entails two paradigms: (1) creating noisy target language training data by translating it from the original source language instances (\ttr) and (2) running inference on noisy source language instances translated from the original target language data (\tts). For token-level tasks, label projection is required for both. We illustrate our pipeline for label projection with \wa in Figure \ref{fig:entityMatch}, comprising the following steps:

\rrparagraph{1. Translating and Pre-Tokenizing (§\ref{sssec:pretokenization})} 
We start from an original source language training instance $S$ in \ttr (or an original target language test instance $T$ in \tts). Next, we translate the instance to the target language for \ttr (or source language for \tts). 
We then pre-tokenize the translated sentence $\hat{T}$ (or $\hat{S}$) for word alignment (i.e., we induce word boundaries).

\rrparagraph{2. Word Alignment} We produce the word alignments between $S$ and $\hat{T}$ for \ttr (or $\hat{S}$ and $T$ for \tts). In the case of \tts, we first gather the label predictions of the \mlm on $\hat{S}$.

\rrparagraph{3. Filtering (§\ref{sssec:filtering})} Based on the word alignment, we collect the labeled spans of tokens that need to be mapped from the source to the target sentence. Optionally, we apply our proposed filtering strategies to identify incomplete alignments between labeled spans of $S$ and the corresponding tokens in $\hat{T}$ for \ttr ($\hat{S}$ and $T$ for \tts). For \ttr, if our filters are met for any labeled span, we discard the training instance $\hat{T}$ (case two in Figure \ref{fig:entityMatch}). For \tts, we modify the procedure as we cannot discard test instances at inference time. Instead, for an incomplete labeled span, we do not map the corresponding labels from $\hat{S}$ to $T$ and assign the default label "O" (case three in Figure \ref{fig:entityMatch}).

\rrparagraph{4. Span Mapping (§\ref{sssec:span_mapping})} Last, we apply our span mapping algorithm to project the labeled spans from $S$ to $\hat{T}$ for \ttr (or $\hat{S}$ to $T$ for \tts).

\subsubsection{Span Mapping}
\label{sssec:span_mapping}
Translation-based \xlt for token-level tasks requires mapping labeled spans of tokens from the source language to the target language sentence.
Instead of simply projecting the token-level labels based on the word alignments produced in step 2 of our pipeline, we propose a more robust, span-based label projection, as illustrated in step 4 of Figure \ref{fig:entityMatch}. 

For \ttr, given a labeled span $L$ in the source sentence $S$ (e.g., $L=\{s_1, \dots, s_4\}$ in Figure \ref{fig:entityMatch}) and the corresponding candidate labeled span $\hat{L}$ in the translated sentence $\hat{T}$ (e.g., $\hat{L}=\{\hat{t}_1, \dots, \hat{t}_4\}$ in Figure \ref{fig:entityMatch})
, we map the labels as follows. We assume a standard BIO scheme in which the first token in a span is labeled with a different tag (B-Tag) than the remaining span (I-Tag). We then map the label of the first token in $L$ to the first token in $\hat{L}$. Next, we map the label of the last token in $L$ to all tokens in $\hat{T}$ between the first and the last token of $\hat{L}$ (inclusively).
The procedure ensures that the mapped labeled spans in $\hat{T}$ are continuous and comply with the BIO scheme.

The span mapping for \tts follows the same procedure. However, instead of mapping the gold labels, from the original source sentence $S$ to the translated target sentence $\hat{T}$, we now map the predictions of the \mlm from the translated source sentence $\hat{S}$ to the original target sentence $T$. 

We also experimented with the simple approach of naively projecting the labels directly based on the word alignments as produced in step 2. This simple projection strategy, however, consistently yielded worse results in our initial experiments: we thus ran the full evaluation using only our span mapping strategy described above. We believe that the simple projection along word alignments performs poorly due to frequent changes in word order between languages.

\subsubsection{Filtering Strategies}
\label{sssec:filtering}
The success of the above span mapping directly depends on the quality of \wa for a concrete language pair, which is affected by (i) the amount of parallel data for the pair used in \wa training, (ii) the amount of monolingual data for the languages in question seen by the \wa's underlying \mlm in pretraining \cite{dou-neubig-2021-word, wang-etal-2022-multilingual}, and (iii) the linguistic proximity between the two languages (and in particular whether they have similar word order). 
To mitigate the impact of imperfect word alignment, we propose several strategies for detecting and filtering alignments of labeled spans with low quality.

\rrparagraph{Complete Source (\comsrc)} We test if all tokens of a span $L = \{s_m,...,s_n\}$ in the source language sentence---$S$ for \ttr and $\hat{S}$ for \tts---have an alignment, i.e., whether the corresponding tokens in $S$ (or $\hat{S}$) are aligned to at least one token in $\hat{T}$ (or $T$). 
Figure \ref{fig:entityMatch} illustrates an example of an incomplete source alignment: in the second and third cases, the token $s_8$ in $S$ (or $\hat{s_8}$ in $\hat{S}$) is not aligned to any token in $\hat{T}$ (or $T$).
We assume that if $L$ is partially unaligned, we miss information from the source language span. Hence, $\hat{L}$ is more likely to be incomplete and thus incorrect. 

\rrparagraph{Complete Target (\comtgt)} The motivation for this filter is analogous to \comsrc: we select only the labeled span alignments for which all span tokens in the target language sentence---$\hat{T}$ for \ttr and $T$ for \tts---are aligned to at least one token in the source language sentence.   
But since we do not have ground truth spans for the target language sentence, we apply the following proxy: we retain only the instances for which the tokens in $\hat{L}$ form a continuous span. Case two and three in Figure \ref{fig:entityMatch} do not satisfy this filter, since $\{\hat{t}_6,\hat{t}_7,\hat{t}_9\}$ (or $\{t_6,t_7,t_9\}$) are discontinuous. We assume that if $\hat{L}$ is partially unaligned, we risk adding additional information to the labeled span in the target language sentence that did not exist in the source language (e.g., by including a target language token distant from the rest of the span).

\rrparagraph{Complete Instance (\comins)} Following \citet{chen-etal-2023-frustratingly}, we verify that the number and type of labeled spans in $S$ and $\hat{T}$ match (e.g., if $S$ has two spans with label \textit{LOC} and one with label \textit{PER} then $\hat{T}$ must also have two \textit{LOC} spans and one \textit{PER} span). This filter can only be applied to \ttr as it needs access to the gold labels of $S$.

\rrparagraph{Restricted Target (\corsch)} 
This filter is specifically designed for \tts.
Preliminary experiments showed that single token labeled spans in $\hat{S}$ are often aligned to multiple discontinuous target tokens in $T$ that have a comparatively large number of unaligned tokens in between. On the one hand, applying our span mapping algorithm without filtering is too coarse-grained---such that it might map a single token labeled span from $\hat{S}$ to a much longer sequence in $T$---on the other hand, applying \comtgt\xspace is too strict---such that it does not map the single token label span from $\hat{S}$ at all. In summary, the former reduces precision while the latter degrades recall. \corsch\xspace aims at the trade-off between no filtering and \comtgt. It maps the single token label span from $\hat{S}$ only to the first token in $\hat{L}$ and assigns the default label "O" to all other tokens. This way, we still map the single token label spans from $\hat{S}$ but limit their length to a single token in $T$.

\rrparagraph{No Filtering (\nofilt)} We additionally compare our filtering strategies against a naive baseline: we omit the filtering step in our label projection pipeline and immediately apply our span mapping algorithm after we obtain the word alignments. 

\subsubsection{Pre-Tokenization}
\label{sssec:pretokenization}
Word alignment for token-level tasks requires the original sentence and the translation to contain word boundaries (i.e., to be pre-tokenized). While the original sentence $S$ (or $T$) is usually given in a pre-tokenized format, we still need to pre-tokenize the translation $\hat{T}$ (or $\hat{S}$). We compare language-agnostic whitespace pre-tokenization (\wstok) against language-specific pre-tokenization (\sptok).\footnote{We provide the details on the tokenizers in App.~\ref{app:expdetails}.} It is worth noting that it is more challenging to pre-tokenize the translated target language sentences $\hat{T}$ in \ttr than the English translations $\hat{S}$ in \tts. 

\subsection{Ensembling \ttr and \tts (\ett)}
\label{ssec:ensemble}
We next detail our proposed translation-based strategy for token-level \xlt that ensembles the predictions of \ttr and \tts \cite{oh-etal-2022-synergy}.
At inference time, the \ttr model produces class logits for each token $t_i$ in the target language sentence $T$. In contrast, the \tts model outputs class logits over each token $\hat{s}_j$ in the translated source language sentence ${\hat{S}}$. 
We then run our \wa-based pipeline for label projection to map the \tts predictions from $\hat{S}$ to $T$. Instead of mapping the predicted class labels for each token, we now map the class logits. Finally, we average the class logits for each token in $T$ obtained from \ttr and \tts. If we cannot project the logits for a token in \tts---because of a missing alignment or because a filter prevents the mapping of a labeled span---we only use the \ttr prediction. Importantly, our proposed ensembling is not restricted to a \wa-based \ttr component and can also be combined with marker-based methods.    

\section{Experimental Setup}
\label{sec:experimental_setup}

\sparagraph{Machine Translation} For translation, we utilize the state-of-the-art massively multilingual NLLB model with 3.3B parameters \cite{nllbteam2022language}. Following prior work \cite{artetxe-etal-2023-revisiting, ebing-glavas-2024-translate}, we decode using beam search with a beam size of $5$.

\rparagraph{Evaluation Tasks} 
We evaluate on two established token classification tasks: named entity recognition and slot labeling. Our experiments span 29 diverse languages, ranging from high-resource languages, represented well in the pretraining corpus of the base \mlm, to low-resource languages, unseen by the \mlm.
We use English as our source language.\footnote{We provide a complete list of target languages in App.~\ref{app:expdetails}.}

\rrparagraph{Named Entity Recognition (NER)} Our evaluation includes 18 of 20 languages from MasakhaNER 2.0 (Masakha) \cite{adelani-etal-2022-masakhaner} supported by the NLLB model used for translation. 
Masakha consists of underserved languages from Sub-Saharan Africa. 
As source data, we use the English training (14k instances) and validation portions (3250 instances) of CoNLL \cite{tjong-kim-sang-de-meulder-2003-introduction}. We add a softmax classifier on top of the \mlm to predict the class for each token.

\rrparagraph{Slot Labeling (SL)} We use the xSID dataset \cite{van-der-goot-etal-2021-masked}, which covers 11 diverse languages and dialects. xSID comprises only evaluation data, so we follow \citet{van-der-goot-etal-2021-masked} and use their publicly released English data for training and validation. The utterances are sourced from the Snips \cite{coucke2018snipsvoiceplatformembedded} and Facebook \cite{schuster-etal-2019-cross-lingual} SL datasets. After deduplication, we end up with over 36k instances for training and 300 for validation.
As for NER, we add a softmax classifier on top of the \mlm. 

\rparagraph{Label Projection}
We compare our \wa-based label projection approach against two state-of-the-art marker-based methods that tag the labeled spans and preserve the tags during translation.  

\rrparagraph{Word Alignment (\wa)}
In our main experiments, we resort to AccAlign \cite{wang-etal-2022-multilingual}, a \wa that is based on the multilingual sentence encoder LaBSE \cite{feng-etal-2022-language}.

\rrparagraph{EasyProject (Easy)} We compare our \wa-based approach against the marker-based label projection method of \newcite{chen-etal-2023-frustratingly}. Before translation, \ep inserts tags ("[", "]") around labeled spans. The MT model is expected to preserve the tags during translation, enabling the reconstruction of the labels afterward. Note that \ep can only be used in \ttr and not in \tts. Let $T$ be the target language sentence at inference time; in \tts, the model will make predictions on its English translation $\hat{S}$; \ep would then insert markers into $\hat
S$ and back-translate to the target language, obtaining $\doublehat{T}$; but $\doublehat{T}$ will generally differ from $T$, which is the actual sentence we need to label.

\rrparagraph{Codec} Our experiments further include Codec \cite{le2024constraineddecodingcrosslinguallabel}, a label projection method that leverages constrained decoding as part of a two-step translation procedure. In the first step, the source sentence is simply translated into the target language (e.g., from English: ``This is New York'' to German: ``Das ist New York''). Then, in the second step, tags are inserted around the labeled spans in the source sentence (English: ``This is [ New York ]''). The marked sentence is fed again as input to the MT model: during decoding, the MT model is now constrained to generate only the tokens from the translation obtained in the first step (``Das'', ``ist'', ``New'', ``York'') or a tag (``['', ``]'').

\rparagraph{Downstream Fine-Tuning}
We use XLM-R Large \cite{conneau-etal-2020-unsupervised} as our base \mlm. For \tts, we also experiment with DeBERTaV3 Large \cite{he2023debertav3improvingdebertausing} and encoder-turned-decoder Llama 3 8B (LLM2Vec) \cite{behnamghader2024llm2vec} as English-centric models. In \ttr, we fine-tune on both the original English data and translated target language data, following \newcite{ebing-glavas-2024-translate} who show that this is better than training only on translations. In \tts, we train the models only on the original English data.
We run all experiments with 3 random seeds and report the mean F\textsubscript{1} score and standard deviation. We provide full training details in Appendix \ref{app:expdetails}.
\begin{table}[t!]
\small
\setlength{\tabcolsep}{1.9pt}
\centering
\begin{tabular}{@{}lccc@{}}
\toprule
                         & Masakha       & xSID          & Avg           \\ \midrule
\multicolumn{4}{c}{\textit{\textbf{Translate-Train}}}                    \\ \midrule
\tnofilt                   & $65.5_{\pm1.2}$          & $82.8_{\pm0.6}$ & $74.2_{\pm1.0}$          \\
\tcomins                   & $65.8_{\pm1.3}$         & $82.8_{\pm0.5}$          & $74.3_{\pm1.0}$ \\
+ \tcomtgt          & $66.0_{\pm1.7}$          & $82.0_{\pm0.7}$          & $74.0_{\pm1.3}$          \\
+ \tcomtgt + \tcomsrc & $66.6_{\pm1.1}$ & $82.0_{\pm0.9}$          & $74.3_{\pm1.0}$ \\ \midrule
\multicolumn{4}{c}{\textit{\textbf{Translate-Test}}}                     \\ \midrule
\tnofilt                   & $46.3_{\pm0.4}$          & $68.2_{\pm0.4}$          & $57.2_{\pm0.4}$          \\
\tcorsch                   & $57.9_{\pm0.5}$          & $74.8_{\pm0.4}$          & $66.4_{\pm0.5}$          \\
+ \tcomtgt          & $57.8_{\pm0.5}$          & 
$74.3_{\pm0.4}$          & $66.0_{\pm0.5}$          \\
+ \tcomsrc          & $58.6_{\pm0.5}$          & 
$73.9_{\pm0.5}$          & $66.2_{\pm0.5}$          \\
+ \tcomtgt + \tcomsrc & $58.0_{\pm0.5}$          & 
$73.2_{\pm0.4}$          & $65.6_{\pm0.5}$          \\ 
\bottomrule
\end{tabular}
\caption{Results on the validation data for \wa-based XLT with various filtering strategies. Results with XLM-R and whitespace pre-tokenization (\wstok).}
\label{tab:filtering}
\vspace{-0.5em}
\end{table}

\section{Results and Discussion}
Starting from the validation portions of our datasets, we assess the impact of low-level design choices related to using \wa for token-level translation-based \xlt (§\ref{sec:pre_results}). Based on these findings, we compare \wa---applying the on average best-performing pre-tokenization and filtering strategies---against two state-of-the-art marker-based methods, \ep and \codec, on the test portions of our datasets (§\ref{sec:main_results}).
Last, we provide further ablations in \S\ref{sec:further_findings}, analyzing the impact of decoder-turned encoder LLMs, \wa, and MT model on the \xlt performance.

\subsection{\wa Design Choices}
\label{sec:pre_results}
\sparagraph{Filtering Strategies}
Our preliminary experiments (Table \ref{tab:filtering}) reveal that filtering has a negligible effect on performance in \ttr: on average 
none of the filtering strategies yields substantial gains. Still, \comins\xspace and \comins+\comtgt+\comsrc\xspace perform the best on average. 
This finding is positive, as searching for an optimal filtering strategy for \ttr is costly: it requires (re-)training language-specific models for every change in the filtering strategy. In stark contrast, applying the \corsch\xspace filter results in a substantial gain (+9.2\% over \nofilt) for \tts. On average, we find \corsch\xspace to be the most successful strategy for \tts. Adding additional filtering (+\comsrc) only results in marginal gains for Masakha. 

\begin{figure}
\centering
\includegraphics{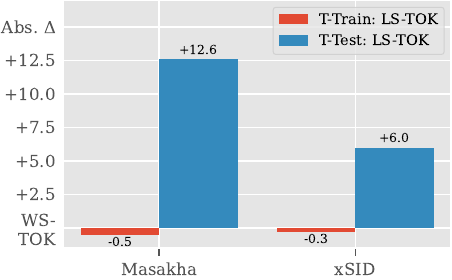}
\caption{Transfer performance with \wa for language-specific pre-tokenization (\sptok) relative to whitespace pre-tokenization (\wstok). Results with XLM-R, \comins+\comtgt+\comsrc\xspace filtering for \ttr and \corsch\xspace filtering for \tts.}
\label{fig:tokenization}
\vspace{-0.5em}
\end{figure}

\rparagraph{Pre-Tokenization}
Figure \ref{fig:tokenization} shows the results for various pre-tokenization approaches. The results mirror the filtering findings: the pre-tokenization strategy has (1) little impact on the \ttr performance---language-agnostic whitespace pre-tokenization (\wstok) is marginally better than language-specific pre-tokenization (\sptok)--- and (2) a substantial impact on \tts---\sptok outperforms \wstok by 12.6\% on Masakha and 6.0\% on xSID. 
Our findings add to prior work \cite{artetxe-etal-2023-revisiting, ebing-glavas-2024-translate}, which showed that \tts is more affected by translation quality than \ttr: our results extend this finding to filtering and pre-tokenization strategies.

\subsection{Main Results}
\label{sec:main_results}
We now compare our optimized \wa-based configurations for \ttr and \tts against the state-of-the-art marker-based label projection methods \ep and \codec. For the remainder of experiments, we apply language-agnostic whitespace pre-tokenization (\wstok) and \comins+\comtgt+\comsrc\xspace filtering for \ttr and language-specific pre-tokenization (\sptok) and \corsch\xspace filtering for \tts.

\rparagraph{Translate-Train} 
We find \ttr, regardless of the label projection strategy (\wa, \ep, or \codec), to substantially outperform zero-shot \xlt with the \mlm (e.g., by 14.2\% on Masakha for \wa-based projection).   
Contrary to the results of prior work that reported translation-based \xlt with \was inferior to \ep \cite{chen-etal-2023-frustratingly} and \codec \cite{le2024constraineddecodingcrosslinguallabel}, we demonstrate that---when reasonably configured---\wa yields competitive performance: on xSID, \wa-based \ttr lags marker-based transfer by less than 1\%; and on Masakha \wa-based transfer even slightly outperforms both \ep and \codec. 

\begin{table}[t!]
\small
\setlength{\tabcolsep}{5.5pt}
\centering
\begin{tabular}{@{}llccc@{}}
\toprule
                &      & Masakha  & xSID  & Avg  \\ \midrule
\multicolumn{5}{c}{\textit{\textbf{Zero-Shot}}}              \\ \midrule
              & X    & $52.9_{\pm1.8}$     & $76.8_{\pm1.4}$     & $64.9_{\pm1.6}$ \\ \midrule
\multicolumn{5}{c}{\textit{\textbf{Translate-Train}}}              \\ \midrule
\ept     & X    & $66.0_{\pm0.9}$     & $83.6_{\pm0.9}$     & $74.8_{\pm0.9}$ \\
\codect           & X    & $66.9_{\pm1.6}$     & $83.4_{\pm0.8}$       & $75.2_{\pm1.3}$    \\
\wa & X    & $67.1_{\pm1.2}$     & $82.7_{\pm0.9}$    & $74.9_{\pm1.0}$ \\ \midrule
\multicolumn{5}{c}{\textit{\textbf{Translate-Test}}}      \\ \midrule
\codect           & X    & $72.0_{\pm0.5}$     & $79.4_{\pm0.3}$       & $75.7_{\pm0.4}$    \\
\codect           & D    & $72.4_{\pm0.4}$     & $79.5_{\pm0.4}$       & $76.0_{\pm0.4}$    \\
\wa  & X    & $72.3_{\pm0.5}$     & $80.2_{\pm0.3}$    & $76.3_{\pm0.4}$ \\
\wa  & D    & $\mathbf{72.7}_{\pm0.4}$     & $80.2_{\pm0.4}$    & $76.5_{\pm0.4}$ \\ \midrule
\multicolumn{5}{c}{\textit{\textbf{Ensemble-Train-Test}}} \\ \midrule
\ept + \wa       & X + D  & $71.7_{\pm0.7}$     & $\mathbf{83.8}_{\pm0.8}$   & $77.7_{\pm0.7}$ \\
\codect + \wa      & X + D  & $72.3_{\pm0.7}$     & $82.8_{\pm0.8}$     & $77.5_{\pm0.7}$    \\
\wa + \wa         & X + D  & $72.6_{\pm0.6}$     & $83.4_{\pm0.9}$   & $\mathbf{78.0}_{\pm0.8}$ \\ \bottomrule
\end{tabular}
\caption{Main results for translation-based XLT for token-level tasks. Results with XLM-R (X) and DeBERTa (D). 
}
\label{tab:main}
\vspace{-0.5em}
\end{table}

\rparagraph{Translate-Test} 
Irrespective of the label projection approach (\wa, Easy, or Codec), \tts outperforms zero-shot \xlt (and \ttr on Masakha). 
%
Again we observe that optimized \wa-based label projection matches and even slightly surpasses the performance of the marker-based \codec. 
This is encouraging because the label projection with \codec---due to its two-step translation procedure---is computationally more expensive (i.e., slower) than \wa. 
Further, we observe that models solely trained on English (i.e., DeBERTa) only offer marginal gains over comparable \mlms (i.e., XLM-R). 
We speculate that this is because NER and SL do not require advanced language understanding abilities and thus the monolingual English ability of an \mlm suffices for these tasks. 

\rparagraph{Ensemble-Train-Test} 
On average, our proposed ensemble \ett improves over \ttr and \tts by 3.1\% and 1.5\%, respectively. We summarize our observations as follows: (i) in scenarios where \ttr performs better than \tts, \ett achieves additional gains over \ttr by leveraging the complementary strengths of \tts; (ii) in scenarios where \ttr performance is worse than \tts, utilizing \ett does not harm because it results in similar performance as \tts.

\begin{table}[t!]
\small
\setlength{\tabcolsep}{3.7pt}
\centering
\begin{tabular}{@{}llccc@{}}
\toprule
    & & Masakha & xSID & Avg  \\ \midrule
\multicolumn{5}{c}{\textit{\textbf{Translate-Test}}}                      \\ \midrule
\twstok & D   & $59.8_{\pm0.3}$   & $73.2_{\pm0.4}$   & $66.6_{\pm0.4}$ \\
\tsptok & D   & $\mathbf{72.7}_{\pm0.4}$    & $80.2_{\pm0.4}$   & $76.5_{\pm0.4}$ \\ \midrule
\multicolumn{5}{c}{\textit{\textbf{Ensemble-Train-Test}}}                 \\ \midrule
WS- + \twstok & X + D  & $72.0_{\pm0.8}$    & $81.2_{\pm1.0}$   & $76.6_{\pm0.9}$ \\
WS- + \tsptok & X + D  & $72.6_{\pm0.6}$    & $\mathbf{83.4}_{\pm0.9}$   & $\mathbf{78.0}_{\pm0.8}$ \\ \bottomrule
\end{tabular}
\caption{Robustness results for \ett utilizing different pre-tokenizations for the \tts component: whitespace (\twstok) and language-specific (\tsptok). Results with XLM-R (X) and DeBERTa (D). 
}
\label{tab:robustness}
\vspace{-0.5em}
\end{table}

\rparagraph{Robustness via Ensembling}
Our preliminary studies on \wa-related low-level design choices (§\ref{sec:pre_results}) revealed notable performance variation, especially for \tts. We now show that our proposed ensemble \ett not only improves performance over both \ttr and \tts but also reduces sensitivity to design details of \wa. 
Table \ref{tab:robustness} compares \tts and \ett with \wa-based label projection for the two pre-tokenization strategies (\wstok and \sptok). For \ett, we modify the pre-tokenization only for the \tts part of the ensemble and keep the \ttr pre-tokenization unchanged. 
We observe that for \tts, \wstok underperforms \sptok by 9.9\% on average. 
\ett, however, almost completely closes the gap between the two, with \wstok trailing \sptok by only 1.4\%, making the choice of the pre-tokenizer much less consequential for the final performance. 

\subsection{Further Findings}
\label{sec:further_findings}
\begin{figure}[t!]
\centering
\includegraphics{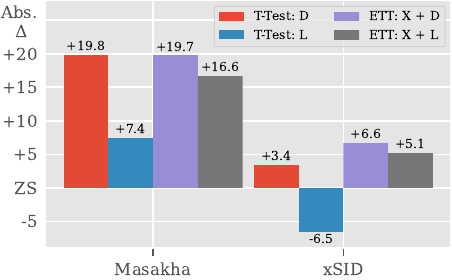}
\caption{Results for translation-based XLT with LLM2Vec (L) vs. DeBERTa (D), relative to zero-shot \xlt performance with XLM-R (X).}
\label{fig:llm2vec}
\vspace{-0.5em}
\end{figure}

\sparagraph{LLMs as Encoders}
Prior work rendered decoder-only LLMs inferior to smaller encoder-only models for token-level tasks \cite{ahuja-etal-2023-mega, le2024constraineddecodingcrosslinguallabel, dukic-snajder-2024-looking}, but more recent efforts suggest that autoregressive LLMs can be post-hoc turned into competitive bidirectional encoders \cite{behnamghader2024llm2vec, wang-etal-2024-improving-text}. 
We thus evaluate a state-of-the-art decoder-turned-encoder Llama 3 8B (LLM2Vec) \cite{behnamghader2024llm2vec} 
in translation-based \xlt for token classification. Following the original work, we add a linear classifier with dropout on top of LLM2Vec, fine-tuning only the classifier. 
Figure \ref{fig:llm2vec} summarizes the results. We observe that much smaller DeBERTa is superior to \llmvec in \tts (i.e., +12.4\% on Masakha and 9.9\% on xSID) and that \tts with LLM2Vec even trails zero-shot transfer with XLM-R on xSID. 
In \ett with an XLM-R-based \ttr component, LLM2Vec becomes much more competitive and lags DeBERTa by only 1.5\% on average, further emphasizing the robustness that our ETT ensembling brings to translation-based \xlt for token-level tasks.

\rparagraph{Choice of Word Aligner}
We next ablate the impact of the \wa model on downstream transfer performance. We compare the widely used Awesome \cite{dou-neubig-2021-word, van-der-goot-etal-2021-masked, chen-etal-2023-frustratingly, le2024constraineddecodingcrosslinguallabel}, based on mBERT, with the more recent AccAlign \cite{wang-etal-2022-multilingual}, which resorts to the multilingual sentence encoder LaBSE. Both \was were released in vanilla and fine-tuned variants.
For the latter, the underlying \mlm is explicitly fine-tuned with word-alignment objectives on parallel data \cite{dou-neubig-2021-word, wang-etal-2022-multilingual}. 
Table \ref{tab:wa} shows the results of the \wa comparison. Without \wa-specific fine-tuning, AccAlign outperforms Awesome by 2.2\% for \ttr and 5.0\% for \tts, respectively. 
The results are mixed w.r.t. explicit \wa fine-tuning: the fine-tuned AccAlign yields virtually no gains in \ttr, but it does bring a small performance boost (1.3\%) in \tts. This is in line with findings from \citet{chen-etal-2023-frustratingly}, who report similar behavior for Awesome. We hypothesize that the limited size and language diversity of \wa fine-tuning limits the generalization to a broader set of (low-resource) languages, as evaluated in our work.

\begin{table}[t!]
\small
\setlength{\tabcolsep}{9pt}
\centering
\begin{tabular}{@{}llccc@{}} 
\toprule
    & Masakha & xSID & Avg  \\ \midrule
\multicolumn{4}{c}{\textit{\textbf{Translate-Train}}} \\ \midrule
 AccAlign       & $67.1_{\pm1.2}$    & $82.7_{\pm0.8}$  & $75.0_{\pm1.0}$ \\
 AccAlgin$_{noft}$ & $66.7_{\pm1.1}$    & $\mathbf{82.9}_{\pm0.5}$  & $74.9_{\pm1.0}$ \\
 Awesome$_{noft}$  & $64.4_{\pm1.3}$    & $79.8_{\pm0.8}$  & $72.7_{\pm1.2}$ \\ \midrule
\multicolumn{4}{c}{\textit{\textbf{Translate-Test}}}  \\ \midrule
 AccAlign       & $\mathbf{72.3}_{\pm0.5}$    & $80.2_{\pm0.3}$  & $\mathbf{76.3}_{\pm0.4}$ \\
 AccAlgin$_{noft}$ & $70.4_{\pm0.5}$    & $79.6_{\pm0.3}$   & $75.0_{\pm0.4}$ \\
 Awesome$_{noft}$  & $65.1_{\pm0.4}$    & $74.8_{\pm0.3}$ & $70.0_{\pm0.4}$ \\ \bottomrule
\end{tabular}
\caption{Comparison of translation-based XLT with different \was; $noft$ denotes vanilla \was without fine-tuning. Results with XLM-R}
\label{tab:wa}
\vspace{-0.5em}
\end{table}

\rparagraph{Translation Quality}
Commercial MT models are typically considered to produce superior translation quality compared to their publicly available counterparts. To evaluate the impact of the MT model on token-level translation-based XLT, we generate translations using Google Translate (GT), which serves as a representative example of a commercial MT model.
We report results for \tts and \ett only (Table \ref{tab:ablation_mt}) as prior work already demonstrated that translation quality has a less pronounced impact on \ttr \cite{artetxe-etal-2023-revisiting, ebing-glavas-2024-translate}. For \tts, we find that GT outperforms NLLB by 1.7\% on average. Nevertheless, the gains obtained by a more powerful MT model are on par with the performance improvements introduced by using our ensemble (\ett) with NLLB only. Additionally, the difference in \ett performance between GT and NLLB is negligible (0.4\%), which once more points to the robustness that \ett brings to \xlt for token classification tasks.

\begin{table}[t!]
\small
\setlength{\tabcolsep}{4.1pt}
\centering
\begin{tabular}{@{}llccc@{}}
\toprule
        &        & Masakha & xSID & Avg  \\ \midrule
\multicolumn{5}{c}{\textit{\textbf{Translate-Test}}}       \\ \midrule
NLLB      & D       & $72.6_{\pm0.4}$    & $80.2_{\pm0.4}$  & $76.4_{\pm0.4}$ \\
GT      & D        & $\mathbf{74.9}_{\pm0.4}$    & $81.2_{\pm0.4}$  & $\mathbf{78.1}_{\pm0.4}$ \\ \midrule
\multicolumn{5}{c}{\textit{\textbf{Ensemble-Train-Test}}}  \\ \midrule
NLLB + NLLB & X + D & $71.9_{\pm0.6}$    & $\mathbf{83.4}_{\pm0.9}$ & $77.7_{\pm0.8}$ \\
NLLB + GT & X + D   & $73.0_{\pm0.6}$    & $83.2_{\pm0.9}$ & $\mathbf{78.1}_{\pm0.8}$ \\ \bottomrule
\end{tabular}
\caption{Results for translation-based XLT using different MT models for \tts: NLLB and Google Translate (GT). Results with XLM-R (X) and DeBERTa (D). 
Scores for Masakha differ as GT does not support all languages.}
\label{tab:ablation_mt}
\vspace{-0.5em}
\end{table}
\section{Conclusion}
In this work, we thoroughly investigated the role of word aligners (\was) in translation-based cross-lingual transfer for token classification tasks. 
Our evaluation on two established benchmarks covering 29 languages, revealed that low-level design decisions related to label projection via \wa can have a substantial effect on translation-based \xlt strategies, in particular translate-test. We then showed that an optimized application of \wa-based label projection can match or even surpass the transfer performance of recent marker-based approaches \cite{chen-etal-2023-frustratingly, le2024constraineddecodingcrosslinguallabel}, contrary to their findings. 
Further, we proposed a more sophisticated \wa-based transfer approach that ensembles predictions of translate-train and translate-test. We demonstrated that the proposed ensemble not only substantially increases transfer performance but also reduces the sensitivity to low-level design decisions of \wa-based label projection. 

\section{Limitations}
We focused on systematically exploring the design choices relevant for translation-based \xlt using \wa. However, our study is limited by the prevalent practice of creating new evaluation datasets by translating the data from an existing high-resource language to the desired (new) language. This applies to xSID and some languages of Masakha. The resulting data may contain distinct characteristics that stem from the translation process often referred to as \textit{translationese}. Prior work \cite{artetxe-etal-2020-translation} stated that translation-based \xlt strategies might lead to the exploitation of translationese, slightly overestimating the true performance. 

For the proposed filtering strategies, we strove for simple and task-agnostic designs. We are aware that the presented filtering strategies are only an excerpt from the endless search space of possible options. However, the contribution of our work does not focus on fully exploring the design space of filtering strategies but demonstrating that filtering strategies for \was may have a substantial impact on the XLT performance. Further, we show that our proposed ensemble successfully mitigates the observed performance variations making suboptimal filtering strategies less influential for the final XLT performance.

Our span mapping approach assumes that the labeled spans projected from the source language sentence $S$ to the target language sentence $\hat{T}$ for \ttr (or $\hat{S}$ to $T$ for \tts) are continuous---such that a single labeled span from the source language sentence may not be split into multiple labeled spans in the target language sentence. This is an (overly) simplifying assumption that might not hold for all pairs of source-to-target (or target-to-source) translations. In our evaluation, the assumption does not hold for some German and Dutch examples in the xSID dataset because the languages allow for sentence final verbs. Therefore, a single slot may be split into discontinuous labeled spans during translation. Nevertheless, our empirical evaluation shows strong results even for German and Dutch indicating that the appearance of discontinuous labeled spans is not too common.

\section*{Acknowledgements}
Simulations were performed with computing resources from Julia 2. Julia 2 was funded as DFG project as “Forschungsgroßgerät nach Art 91b GG” under INST 93/1145-1 FUGG"


\bibliography{anthology,custom}

\appendix

\newpage
\section{Translation Data}

For Masakha \cite{adelani-etal-2022-masakhaner} and xSID \cite{van-der-goot-etal-2021-masked}, we concatenated the pre-tokenized input on whitespace before translation. We deviate from this for the Chinese data in xSID, where we merge Chinese tokens without whitespace. Additionally, the dialect \textit{South Tyrol} (de-st) in xSID is not supported by NLLB. We translate the dialect pretending it to be German (i.e., using the German language code) as it is closely related to the latter. Further, the Serbian (sr) data in xSID is written in Latin script, whereas NLLB was only trained in Serbian Cyrillic script. We accessed all datasets through the Hugging Face library and ensured compliance with the licenses.

\section{Word Alignment}

For our main experiments, we use the neural word aligner AccAlign \cite{wang-etal-2022-multilingual}, accessed through the following repository: \href{https://github.com/sufenlp/AccAlign}{https://github.com/sufenlp/AccAlign}. Additionally, we employ Awesome \cite{dou-neubig-2021-word} with the code provided in the following repository: \href{https://github.com/neulab/awesome-align}{https://github.com/neulab/awesome-align}. We follow the hyperparameter configuration proposed by the authors. We ensure compliance with the license for Awesome (BSD 3-Clause). We could not find licensing information for AccAlign.  

\section{Easy}

The code and data of \ep is released under the MIT license. We used the publicly released data for Masakha. For xSID, we produced our own translated data by adopting the existing code. We followed their implementation for Masakha closely. \ep \cite{chen-etal-2023-frustratingly} requires fine-tuning NLLB for preserving inserted tags (i.e., preserving "[" and "]" around labeled spans). Hence, we leverage their publicly released 3.3B parameter checkpoint from \citet{chen-etal-2023-frustratingly} for translation. We accessed it through the Hugging Face library. 

\section{Codec}
\label{ssec:codec}

The authors of Codec did not release the translated data but published the source code instead. We created our own translated data for Masakha following their implementation. Further, we extended their implementation to produce the translated data for xSID. We adhered to the hyperparameters in their repository and followed the existing implementation closely. The translations for Codec are obtained using vanilla (i.e., non fine-tuned) NLLB. However, the constrained decoding (i.e., inserting the tags post-translation) requires a fine-tuned NLLB that is able to preserve/insert tags. Therefore, for constrained decoding, we follow \citet{le2024constraineddecodingcrosslinguallabel} using the fine-tuned 600M parameter version of NLLB released by \citet{chen-etal-2023-frustratingly}. We could not find licensing information for Codec.

\section{T-Projection}
\label{app:tproj}
T-Projection (T-Proj) \cite{garcia-ferrero-etal-2023-projection} solves the task of identifying the labeled spans in the translated target language sentence as a sequence-to-sequence problem. In the first step, the translated target language sentence is obtained by machine-translating it from the original source sentence. Next, the translated sentence and the type of the labeled span (i.e., LOC in NER) are fed to a sequence-to-sequence model that generates multiple candidates for the corresponding labeled span in the translated sentence. Finally, the best candidate is chosen by computing a translation quality metric between the labeled span from the original source language sentence and the candidates produced by the sequence-to-sequence model. For our preliminary results
shown in Appendix \ref{app:ablation_proj}, we reuse the data for Masakha released by \citet{garcia-ferrero-etal-2023-projection} in the following repository: \href{https://github.com/ikergarcia1996/T-Projection}{https://github.com/ikergarcia1996/T-Projection}. We ensure compliance with the T-Proj license (Apache-2.0).

\section{CLaP}
\label{app:clap}
CLaP \cite{parekh-etal-2024-contextual} leverages \textit{contextualized translation} to map the labeled spans from the original to the translated sentence. First, the original sentence is machine-translated to the target language. Second, an LLM is prompted to perform contextualized translation: given the translated target language sentence and the labeled spans of the original sentence, the LLM is prompted to output the labeled spans in the translated target language sentence. For our preliminary results
shown in Appendix \ref{app:ablation_proj}, we reuse the implementation of \citet{parekh-etal-2024-contextual} provided in the following repository: \href{https://github.com/PlusLabNLP/CLaP}{https://github.com/PlusLabNLP/CLaP}. Furthermore, we follow their experimental setup closely. We use the text completion version of Llama-2 \cite{touvron2023llama2openfoundation} with 13B parameters. Together with their prompt template, we provide two in-context examples per language. We create these examples with GPT-4o mini \cite{openai2024gpt4technicalreport} using the same prompt template used for CLaP. We manually validate the correctness of the translated target language entities by back-translating them with GT. We could not find licensing information for CLaP.

\section{Detailed Experimental Setup}
\label{app:expdetails}
We train both tasks (NER and SL) for 10 epochs using an effective batch size of 32. In case we can not fit the desired batch size, we utilize gradient accumulation. The learning rate is set to $1e^{-5}$ with a weight decay of 0.01. We implement a linear schedule of 10\% warm-up and employ mixed precision. For the LLM2Vec experiments, we deviate from this setting as we only fine-tune the classifier. Following \citet{behnamghader2024llm2vec}, we set the learning rate to $5e^{-4}$. We evaluate models at the last checkpoint of training. We use the seqeval F1 implementation accessed through the Hugging Face library. Further, we access our downstream models---XLM-RoBERTa Large, DeBERTaV3 Large, and LLM2Vec Llama 3 8B Instruct MNTP---through the Hugging Face library. All translations were run on a single A100 with 40GB VRAM, and all downstream training and evaluation runs were completed on a single V100 with 32GB VRAM. We estimate the GPU time to 4000 hours across all translations and downstream fine-tunings.

\rparagraph{Languages}

\rrparagraph{MasakhaNER2.0} Our experiments cover 18 out of 20 languages that are supported by NLLB. Note that Google Translate (GT) does not support all 18 languages. Following, we mark the 11 languages that are supported by GT with an additional asterisk:  Bambara (bam)*, Ewé (ewe)*, Fon (fon), Hausa (hau)*, Igbo (ibo)*, Kinyarwanda (kin)*, Luganda (lug), Luo (luo), Mossi (most), Chichewa (nya), chiShona (sna)*, Kiswahili (saw)*, Setswana (tsn), Akan/Twi (twi)*, Wolof (wol), isiXhosa (xho)*, Yorùrbá (yor)*, and isiZulu (zul)*. 

\rrparagraph{xSID} We evaluate 11 languages all covered by NLLB and GT: Arabic (ar), Danish (da), German (de), South-Tyrolean (de-st), Indonesian (id), Italian (it), Kazakh (kk), Dutch (nl), Serbian (sr), Turkish (tr), and Chinese (zh). Following \citet{razumovskaia-etal-2023-transfer}, we excluded Japanese from the evaluation because it only has half of the validation and test instances and spans only a fraction of entities compared to the other languages.

\rparagraph{Filtering Strategy}
We use a greedy approach to explore the various design options (§\ref{sec:pre_results}). We start with the selection of the filtering strategy, followed by our pre-tokenization experiments. For the exploration of the filtering strategy, we apply whitespace pre-tokenization (\wstok). 

\rparagraph{Pre-Tokenization}
For the per-tokenization experiments, we filter the translated training data based on \comins+\comsrc+\comtgt. For \tts, we apply \corsch. Language-specific tokenization (\sptok) is done with the MosesTokenizer from the Sacremoses library (\href{https://github.com/hplt-project/sacremoses}{https://github.com/hplt-project/sacremoses}), except for Chinese, where we use jieba (\href{https://github.com/fxsjy/jieba}{https://github.com/fxsjy/jieba}). Both tokenizers are released under the MIT license.

\rparagraph{Main Results and Further Findings}
As suggested by the findings of our preliminary experiments, we apply whitespace pre-tokenization (\wstok) for \ttr, except for Chinese in xSID, where we use language-specific tokenization (\sptok). For \tts, we use language-specific pre-tokenization (\sptok). We utilize the same filtering as for the pre-tokenization experiments: \comins+\comsrc+\comtgt\xspace for \ttr and \corsch\xspace for \tts.

\rparagraph{Ensembling with Codec as T-Test Component}
In our main results (see Table \ref{tab:main}), we report \ett only with a \wa-based \tts component while generally, the ensembling approach is not limited to that. For the \tts component, an explicit mapping between the $m$ labeled spans from the translated sentence to the $n$ labeled spans in the original sentence is required such that the logits can be projected from the translated to the original sentence. The current implementation of Codec does not explicitly provide this mapping since it only outputs the labeled translated sentence. Obtaining the mapping would require a non-trivial extension of the implementation of Codec. Nevertheless, we argue that if the extension is implemented, there is no reason why \ett with a Codec \tts component should perform worse than with a \wa\xspace \tts component. However, we do not expect it to perform better either since the individual \ttr and \tts performance of Codec and our \wa-based approach is comparable.

\begin{table*}[t!]
\section{Detailed Results: Filtering Strategies}
\label{app:detailed_filtering}
\scriptsize
\centering
\setlength{\tabcolsep}{2.6pt}
\begin{tabular}{@{}lccccccccccccccccccc@{}}
\toprule
                         & bam  & ewe  & fon  & hau  & ibo  & kin  & lug  & luo  & mos  & nya  & sna  & swa  & tsn  & twi  & wol  & xho  & yor  & zul  & Avg  \\
\midrule
\multicolumn{20}{c}{\textit{\textbf{Translate-Train}}}                                                                                                        \\
\midrule
\tnofilt                         & 43.8 & 78.4 & 72.9 & 66.7 & 62.9 & 70.2 & 79.6 & 69.7 & 58.9 & 65.2 & 72.5 & 84.4 & 69.1 & 57.3 & 63.3 & 63.2 & 33.7 & 67.1 & 65.5 \\
\tcomins                         & 45.5 & 78.4 & 76.0 & 66.5 & 62.8 & 69.5 & 79.4 & 70.2 & 59.9 & 66.1 & 73.1 & 84.1 & 69.0 & 56.6 & 65.8 & 62.8 & 32.9 & 66.5 & 65.8 \\
\tcomins + \tcomtgt                         & 50.9 & 77.8 & 76.0 & 66.5 & 62.5 & 69.0 & 80.4 & 70.4 & 60.4 & 65.4 & 73.0 & 83.6 & 68.9 & 55.9 & 67.0 & 63.1 & 33.2 & 64.7 & 66.0 \\
\tcomins + \tcomtgt + \tcomsrc                         & 51.2 & 78.8 & 76.5 & 66.5 & 63.7 & 69.7 & 80.0 & 70.0 & 60.3 & 66.2 & 73.1 & 83.6 & 68.2 & 60.8 & 67.3 & 63.4 & 32.6 & 66.3 & 66.6 \\
\midrule
\multicolumn{20}{c}{\textit{\textbf{Translate-Test}}}                                                                                                         \\ 
\midrule
\tnofilt                  & 26.2 & 49.0 & 40.0 & 55.5 & 43.3 & 55.7 & 58.6 & 52.9 & 36.6 & 51.9 & 52.6 & 55.3 & 52.6 & 54.3 & 42.9 & 30.9 & 29.0 & 45.7 & 46.3 \\
\tcorsch                 & 37.6 & 68.9 & 58.0 & 61.7 & 55.5 & 67.1 & 70.9 & 61.6 & 46.4 & 64.7 & 60.9 & 67.4 & 63.2 & 61.6 & 55.6 & 43.1 & 43.0 & 55.5 & 57.9 \\
\tcomtgt                  & 32.5 & 52.0 & 47.2 & 56.9 & 44.5 & 56.8 & 59.7 & 55.2 & 40.6 & 52.9 & 53.0 & 55.7 & 54.1 & 59.6 & 46.0 & 31.5 & 32.8 & 46.2 & 48.7 \\
\tcomsrc                  & 28.9 & 47.7 & 41.5 & 56.3 & 43.4 & 55.6 & 58.8 & 54.0 & 35.1 & 52.1 & 52.7 & 54.5 & 51.8 & 56.4 & 44.2 & 30.8 & 28.7 & 45.9 & 46.6 \\
\tcomtgt + \tcomsrc          & 34.4 & 50.5 & 47.9 & 57.4 & 44.6 & 56.7 & 59.8 & 55.7 & 38.8 & 53.0 & 53.0 & 54.8 & 53.1 & 61.1 & 46.5 & 31.4 & 32.7 & 46.4 & 48.8 \\
\tcorsch + \tcomtgt          & 36.8 & 67.7 & 52.4 & 62.9 & 55.9 & 66.8 & 71.6 & 62.5 & 45.0 & 65.0 & 61.0 & 68.0 & 64.2 & 61.1 & 57.0 & 43.6 & 43.4 & 55.6 & 57.8 \\
\tcorsch + \tcomsrc         & 41.7 & 68.2 & 60.5 & 62.6 & 55.8 & 67.0 & 71.2 & 63.0 & 45.4 & 65.0 & 61.1 & 66.8 & 62.6 & 64.0 & 57.5 & 43.1 & 42.9 & 55.7 & 58.6 \\
\tcorsch + \tcomtgt + \tcomsrc & 39.1 & 66.6 & 53.2 & 63.5 & 56.1 & 66.7 & 71.7 & 63.1 & 43.4 & 65.1 & 61.1 & 67.2 & 63.3 & 62.7 & 57.8 & 43.5 & 43.4 & 55.8 & 58.0 \\ \bottomrule
\end{tabular}
\caption{Results for translation-based XLT evaluated on the Masakha validation data utilizing different filtering strategies. We use XLM-R.}
\end{table*}

\begin{table*}[t!]
\small
\centering
\setlength{\tabcolsep}{4.5pt}
\begin{tabular}{@{}lcccccccccccc@{}}
\toprule
 & ar   & da   & de   & de-st & id   & it   & kk   & nl   & sr   & tr   & zh   & Avg  \\ \midrule
\multicolumn{13}{c}{\textit{\textbf{Translate-Train}}}                                \\ \midrule
\tnofilt & 85.4 & 81.8 & 88.3 & 60.3  & 86.2 & 89.8 & 70.5 & 94.1 & 85.7 & 85.9 & -    & 82.8 \\
\tcomins & 85.1 & 82.4 & 88.7 & 58.3  & 86.0 & 90.1 & 70.7 & 93.3 & 84.6 & 88.0 & -    & 82.7 \\
\tcomins + \tcomtgt & 84.9 & 81.9 & 87.8 & 59.3  & 85.3 & 86.7 & 69.8 & 90.9 & 86.6 & 86.6 & -    & 82.0 \\
\tcomins + \tcomtgt + \tcomsrc  & 86.3 & 81.3 & 86.7 & 58.8  & 85.0 & 88.1 & 69.6 & 91.3 & 86.2 & 86.5 & -    & 82.0 \\
\midrule
\multicolumn{13}{c}{\textit{\textbf{Translate-Test}}}                                 \\ \midrule
\tnofilt& 68.1& 75.6& 74.4& 51.9 & 73.1& 76.6& 51.7& 80.2& 69.9& 66.6& 61.6& 68.2\\

\tcorsch                  & 73.6& 78.1& 82.7& 57.3 & 73.5& 83.4& 61.7& 86.4& 75.9& 75.1& 75.4& 74.8\\

\tcomtgt& 68.4& 75.5& 79.4& 53.8 & 74.3& 75.5& 59.8& 79.8& 70.6& 74.9& 74.6& 71.5\\

\tcomsrc& 67.2& 74.3& 73.4& 51.7 & 71.5& 75.5& 50.4& 78.9& 68.2& 65.2& 61.0& 67.0\\

\tcomtgt + \tcomsrc           & 67.4& 74.3& 78.8& 53.4 & 72.8& 74.8& 59.0& 78.9& 68.7& 72.8& 73.5& 70.4\\

\tcorsch + \tcomtgt           & 74.1& 77.9& 81.7& 57.6 & 74.0& 82.1& 60.1& 84.6& 76.3& 73.8& 74.9& 74.3\\

\tcorsch + \tcomsrc          & 72.9& 77.0& 82.0& 57.2 & 71.9& 82.5& 60.8& 85.2& 74.3& 73.9& 75.1& 73.9\\

\tcorsch + \tcomtgt + \tcomsrc  & 73.3& 76.7& 81.1& 57.3 & 72.4& 81.6& 59.3& 83.8& 74.5& 71.7& 73.8& 73.2\\ \bottomrule
\end{tabular}
\caption{Results for translation-based XLT evaluated on the xSID validation data utilizing different filtering strategies. We use XLM-R. For \ttr, we excluded Chinese (zh) since experiments were run with whitespace pre-tokenization (\wstok).}
\end{table*}

\begin{table*}[t!]
\section{Filtering Strategies: Recovered Instances for Translate-Train}
\scriptsize
\centering
\setlength{\tabcolsep}{2.6pt}
\begin{tabular}{@{}lccccccccccccccccccc@{}}
\toprule
 & bam  & ewe  & fon  & hau  & ibo  & kin  & lug  & luo  & mos  & nya  & sna  & swa  & tsn  & twi  & wol  & xho  & yor  & zul  & Avg  \\ \midrule
\tcomins & 97.5 & 98.4 & 95.0 & 99.5 & 99.2 & 98.4 & 98.5 & 98.3 & 93.7 & 99.3 & 99.5 & 99.7 & 99.2 & 98.5 & 96.0 & 99.0 & 98.7 & 99.3 & 98.2 \\
\tcomins + \tcomtgt & 96.5 & 97.0 & 94.1 & 96.7 & 98.0 & 92.6 & 97.2 & 97.0 & 92.3 & 96.8 & 99.3 & 94.5 & 96.5 & 97.7 & 94.3 & 98.6 & 97.2 & 98.9 & 96.4 \\
\tcomins + \tcomtgt + \tcomsrc & 93.6 & 94.7 & 92.2 & 95.0 & 97.3 & 91.6 & 95.9 & 95.6 & 89.2 & 96.1 & 98.6 & 93.9 & 95.7 & 95.8 & 91.6 & 97.4 & 96.0 & 97.3 & 94.9 \\ \bottomrule
\end{tabular}
\caption{Relative number of recovered instances on the Masakha validation data utilizing different filtering strategies.} 
\end{table*}

\begin{table*}[t!]
\small
\centering
\setlength{\tabcolsep}{4.7pt}
\begin{tabular}{@{}lcccccccccccc@{}}
\toprule
 & ar   & da   & de   & de-st & id   & it   & kk   & nl   & sr   & tr   & zh & Avg  \\ \midrule
\tcomins & 93.5 & 97.9 & 96.4 & 96.4  & 98.2 & 97.8 & 90.6 & 96.6 & 97.2 & 92.0 & -  & 95.7 \\
\tcomins + \tcomtgt & 85.9 & 92.4 & 85.4 & 85.4  & 92.1 & 78.6 & 81.6 & 84.4 & 91.5 & 87.0 & -  & 86.4 \\
\tcomins + \tcomtgt + \tcomsrc & 70.8 & 82.3 & 78.5 & 78.5  & 83.1 & 71.0 & 65.9 & 79.5 & 75.7 & 71.9 & -  & 75.7 \\ \bottomrule
\end{tabular}
\caption{Relative number of recovered instances on the xSID validation data utilizing different filtering strategies. We excluded Chinese (zh) since filtering was run with whitespace pre-tokenization (\wstok).}
\end{table*}

\begin{table*}[t!]
\section{Filtering Strategies: Mapped Labeled Spans for Translate-Test}
\scriptsize
\centering
\setlength{\tabcolsep}{1pt}
\begin{tabular}{@{}lccccccccccccccccccc@{}}
\toprule
 & bam   & ewe   & fon   & hau   & ibo   & kin   & lug   & luo   & mos   & nya   & sna   & swa   & tsn   & twi   & wol   & xho   & yor   & zul   & Avg   \\ \midrule
\tcorsch & 101.0 & 100.5 & 100.3 & 100.1 & 100.2 & 100.0 & 100.1 & 100.1 & 100.4 & 100.1 & 100.2 & 100.1 & 100.2 & 100.1 & 100.2 & 100.1 & 101.4 & 100.3 & 100.3 \\
\tcorsch + \tcomtgt & 70.5  & 87.4  & 71.8  & 92.1  & 94.7  & 94.1  & 95.3  & 92.7  & 79.8  & 96.7  & 99.1  & 88.6  & 91.4  & 81.7  & 86.6  & 97.9  & 86.8  & 98.6  & 89.2  \\
\tcorsch + \tcomsrc & 82.8  & 94.7  & 90.7  & 97.0  & 98.7  & 99.4  & 98.9  & 92.9  & 89.6  & 99.0  & 98.6  & 97.0  & 96.6  & 91.6  & 90.9  & 99.0  & 98.0  & 99.6  & 95.3  \\
\tcorsch + \tcomtgt + \tcomsrc & 60.1  & 82.7  & 67.2  & 90.2  & 93.4  & 93.7  & 94.5  & 87.6  & 72.3  & 95.9  & 97.8  & 86.6  & 89.2  & 76.4  & 81.7  & 97.1  & 84.1  & 97.8  & 86.0  \\ \bottomrule
\end{tabular}
\caption{Fraction of mapped labeled spans for the Masakha validation data relative to \nofilt\xspace utilizing different filtering strategies. Filtering strategies might prevent interference between labeled spans of the same instance resulting in more labeled spans than in \nofilt.}
\end{table*}

\begin{table*}[t!]
\small
\centering
\setlength{\tabcolsep}{2.4pt}
\begin{tabular}{@{}lcccccccccccc@{}}
\toprule
 & ar    & da    & de    & de-st & id    & it    & kk    & nl    & sr    & tr    & zh    & Avg   \\ \midrule
\tcorsch  & 100.2 & 100.2 & 100.4 & 103.7 & 100.6 & 100.2 & 109.2 & 100.9 & 100.2 & 108.6 & 112.4 & 102.4 \\
\tcorsch + \tcomtgt & 96.8  & 97.3  & 93.0  & 93.0  & 96.5  & 94.8  & 95.6  & 95.0  & 95.3  & 95.8  & 98.2  & 95.3  \\
\tcorsch + \tcomsrc & 93.7  & 95.3  & 96.3  & 98.7  & 96.4  & 95.2  & 103.8 & 97.4  & 95.2  & 102.0 & 108.2 & 97.4  \\
\tcorsch + \tcomtgt + \tcomsrc & 90.8  & 93.1  & 90.3  & 90.0  & 92.8  & 91.3  & 92.1  & 93.2  & 90.7  & 90.4  & 94.2  & 91.5  \\ \bottomrule
\end{tabular}
\caption{Fraction of mapped labeled spans for the xSID validation data relative to \nofilt\xspace utilizing different filtering strategies. Filtering strategies might prevent interference between labeled spans of the same instance resulting in more labeled spans than in \nofilt.}
\end{table*}

\begin{table*}[t!]
\section{Detailed Results: Pre-Tokenization}
\label{app:detailed_tokenization}
\small
\centering
\setlength{\tabcolsep}{3.2pt}
\begin{tabular}{@{}lccccccccccccccccccc@{}}
\toprule
 & bam  & ewe  & fon  & hau  & ibo  & kin  & lug  & luo  & mos  & nya  & sna  & swa  & tsn  & twi  & wol  & xho  & yor  & zul  & Avg  \\ \midrule
\multicolumn{20}{c}{\textit{\textbf{Translate-Train}}}                                                                                \\ \midrule
\twstok & 51.2 & 78.8 & 76.5 & 66.5 & 63.7 & 69.7 & 80.0 & 70.0 & 60.3 & 66.2 & 73.1 & 83.6 & 68.2 & 60.8 & 67.3 & 63.4 & 32.6 & 66.3 & 66.6 \\
\tsptok & 49.4 & 77.9 & 75.3 & 66.7 & 58.4 & 68.8 & 80.2 & 70.7 & 61.3 & 65.9 & 73.7 & 83.4 & 69.8 & 59.2 & 66.9 & 63.8 & 32.3 & 66.8 & 66.1 \\ \midrule
\multicolumn{20}{c}{\textit{\textbf{Translate-Test}}}                                                                                 \\ \midrule
\twstok & 37.6 & 68.9 & 58.0 & 61.7 & 55.5 & 67.1 & 70.9 & 61.6 & 46.4 & 64.7 & 60.9 & 67.4 & 63.2 & 61.6 & 55.6 & 43.1 & 43.0 & 55.5 & 57.9 \\
\tsptok & 50.8 & 82.5 & 74.8 & 67.0 & 71.5 & 78.8 & 86.8 & 76.2 & 55.2 & 75.8 & 82.0 & 81.9 & 76.1 & 68.8 & 66.1 & 62.9 & 54.7 & 69.4 & 71.2 \\ \bottomrule
\end{tabular}
\caption{Results for translation-based XLT evaluated on the Masakha validation data utilizing different pre-tokenization strategies. We use XLM-R.}
\end{table*}

\begin{table*}[]
\small
\centering
\setlength{\tabcolsep}{9.3pt}
\begin{tabular}{@{}lcccccccccccc@{}}
\toprule
 & ar   & da   & de   & de-st & id   & it   & kk   & nl   & sr   & tr   & zh    & Avg  \\ \midrule
\multicolumn{13}{c}{\textit{\textbf{Translate-Train}}}                                 \\ \midrule
\twstok & 86.3 & 81.3 & 86.7 & 58.8  & 85.0 & 88.1 & 69.6 & 91.3 & 86.2 & 86.5 & -     & 82.0 \\
\tsptok & 86.4 & 81.4 & 87.8 & 57.4  & 85.8 & 87.3 & 69.7 & 90.8 & 83.8 & 86.3 & 86.2* & 81.6 \\ \midrule
\multicolumn{13}{c}{\textit{\textbf{Translate-Test}}}                                  \\ \midrule
\twstok & 73.6 & 78.1 & 82.7 & 57.3 & 73.5 & 83.4 & 61.7 & 86.4 & 75.9 & 75.1 & 75.4 & 74.8 \\
\tsptok & 79.1 & 81.3 & 89.7 & 62.7 & 78.7 & 89.6 & 69.9 & 92.9 & 80.4 & 81.7 & 82.7 & 80.8 \\ \bottomrule
\end{tabular}
\caption{Results for translation-based XLT evaluated on the xSID validation data utilizing different pre-tokenization strategies. We use XLM-R. Results marked with * are excluded from the average.}
\end{table*}

\begin{table*}[t!]
\section{Detailed Results: Main Results and Further Findings}
\label{app:detailed_main}
\scriptsize
\centering
\setlength{\tabcolsep}{1.6pt}
\begin{tabular}{@{}llllccccccccccccccccccc@{}}
\toprule
             &       &               &      & bam  & ewe  & fon  & hau  & ibo  & kin  & lug  & luo  & mos  & nya  & sna  & swa  & tsn  & twi  & wol  & xho  & yor  & zul  & Avg  \\ \midrule
\zs           &       &               &      & 43.4 & 72.8 & 61.0 & 73.5 & 49.9 & 46.3 & 64.9 & 55.0 & 56.1 & 51.1 & 34.4 & 88.1 & 51.5 & 49.5 & 56.2 & 22.2 & 35.1 & 41.5 & 52.9 \\ \midrule
\multicolumn{23}{c}{\textit{\textbf{Translate-Train}}}                                                                                                                           \\ \midrule
AccAlign          & X     & \twstok        & NLLB & 49.2 & 74.1 & 72.1 & 73.1 & 72.2 & 58.6 & 76.4 & 63.5 & 58.2 & 66.2 & 70.9 & 83.2 & 76.5 & 64.1 & 63.9 & 69.6 & 40.2 & 75.6 & 67.1 \\
AccAlign$^\dag$     & X     & \twstok        & NLLB & 53.3 & 74.0 & 71.3 & 73.6 & 71.0 & 58.8 & 75.5 & 64.2 & 55.6 & 67.6 & 68.7 & 83.6 & 76.1 & 64.5 & 61.9 & 69.4 & 39.3 & 72.6 & 66.7 \\
Awesome$^\dag$ & X     & \twstok        & NLLB & 51.3 & 73.8 & 65.6 & 73.6 & 70.0 & 56.7 & 74.4 & 64.6 & 50.8 & 67.3 & 68.4 & 82.2 & 75.3 & 62.4 & 58.9 & 61.0 & 38.4 & 64.7 & 64.4 \\
Easy         & X     & -        & NLLB & 54.2 & 75.4 & 71.1 & 73.0 & 64.6 & 66.3 & 77.5 & 63.8 & 51.3 & 68.3 & 57.2 & 84.1 & 74.7 & 63.7 & 63.3 & 71.3 & 37.0 & 70.6 & 66.0 \\
Codec        & X     & -        & NLLB & 51.2 & 74.1 & 68.9 & 73.4 & 65.5 & 64.7 & 75.4 & 64.7 & 53.9 & 68.3 & 70.9 & 84.2 & 73.5 & 65.2 & 65.6 & 70.2 & 39.4 & 75.3 & 66.9 \\ \midrule
\multicolumn{23}{c}{\textit{\textbf{Translate-Test}}}                                                                                                                            \\ \midrule
AccAlign          & X     & \tsptok        & NLLB & 54.7 & 79.1 & 72.4 & 74.0 & 73.9 & 70.4 & 83.8 & 73.3 & 52.7 & 78.6 & 81.2 & 83.1 & 79.2 & 70.0 & 66.1 & 72.8 & 57.8 & 78.5 & 72.3  \\
AccAlign$^\dag$     & X     & \tsptok        & NLLB & 54.1 & 76.3 & 69.1 & 73.4 & 72.4 & 69.6 & 82.7 & 71.4 & 48.2 & 77.6 & 80.0 & 81.7 & 79.4 & 70.5 & 62.8 & 71.1 & 49.4 & 77.0 & 70.4  \\
Awesome$^\dag$ & X     & \tsptok        & NLLB & 46.3 & 72.4 & 58.6 & 69.4 & 75.3 & 65.0 & 81.6 & 72.1 & 42.4 & 78.4 & 66.3 & 80.0 & 78.3 & 68.9 & 53.4 & 52.1 & 47.7 & 64.3 & 65.1  \\
AccAlign          & D     & \tsptok        & NLLB & 54.3 & 79.2 & 73.3 & 74.5 & 75.1 & 71.2 & 84.0 & 75.0 & 52.5 & 79.1 & 81.6 & 83.7 & 79.3 & 70.2 & 66.3 & 72.9 & 57.6 & 78.2 & 72.7  \\
AccAlign          & X     & \twstok        & NLLB & 44.6 & 69.6 & 55.8 & 63.3 & 59.1 & 58.4 & 72.9 & 60.2 & 40.7 & 68.2 & 60.8 & 68.3 & 63.7 & 60.9 & 55.2 & 61.8 & 46.2 & 62.1 & 59.5  \\
AccAlign          & D     & \twstok        & NLLB & 44.3 & 69.7 & 56.2 & 63.8 & 59.9 & 58.9 & 72.6 & 61.6 & 40.4 & 68.7 & 61.4 & 68.8 & 64.3 & 61.3 & 55.8 & 61.7 & 46.2 & 61.7 & 59.8   \\
AccAlign          & X     & \tsptok        & GT   & 60.9 & 79.3 & -    & 73.4 & 78.0 & 71.7 & -    & -    & -    & -    & 83.4 & 85.2 & -    & 71.3 & -    & 75.0 & 63.7 & 78.2 & -    \\
AccAlign          & D     & \tsptok        & GT   & 60.2 & 79.4 & -    & 74.1 & 79.1 & 72.3 & -    & -    & -    & -    & 83.9 & 85.9 & -    & 72.9 & -    & 75.1 & 62.2 & 78.4 & -    \\
AccAlign          & ShL   & \tsptok        & NLLB & 50.0 & 66.9 & 62.6 & 61.9 & 59.1 & 58.7 & 67.1 & 61.6 & 43.5 & 70.7 & 64.4 & 71.8 & 70.0 & 59.4 & 57.3 & 59.4 & 44.9 & 60.8 & 60.6   \\
AccAlign          & L     & \tsptok        & NLLB & 46.2 & 68.1 & 61.0 & 62.1 & 58.0 & 56.9 & 69.9 & 60.9 & 43.0 & 69.0 & 67.8 & 72.5 & 66.7 & 56.4 & 56.9 & 59.3 & 45.5 & 65.5 & 60.3   \\
Codec        & X     & -        & NLLB & 54.5 & 78.8 & 67.4 & 72.9 & 72.8 & 77.6 & 83.6 & 72.8 & 49.4 & 78.1 & 79.3 & 82.2 & 79.2 & 72.5 & 67.3 & 72.5 & 58.4 & 77.1 & 72.0 \\
Codec        & D     & -        & NLLB & 54.3 & 79.1 & 68.0 & 73.3 & 73.9 & 78.2 & 83.5 & 74.2 & 48.8 & 79.0 & 79.8 & 82.9 & 79.3 & 73.1 & 67.8 & 72.6 & 58.0 & 77.0 & 72.4 \\ \midrule
\multicolumn{23}{c}{\textit{\textbf{Ensembling-Translate-Train}}}                                                                                                                \\ \midrule
AccAlign + AccAlign          & X + X   & WS- + \twstok & NLLB & 57.0 & 79.1 & 74.3 & 72.6 & 77.6 & 63.5 & 81.8 & 69.8 & 59.4 & 74.9 & 75.6 & 83.8 & 78.4 & 66.3 & 67.6 & 72.0 & 52.6 & 78.7 & 71.4  \\
AccAlign + AccAlign          & X + X   & WS- + \tsptok & NLLB & 57.3 & 78.5 & 75.7 & 72.8 & 79.0 & 64.1 & 82.5 & 70.3 & 60.2 & 75.3 & 77.2 & 84.3 & 78.9 & 66.7 & 68.0 & 72.4 & 53.5 & 78.8 & 72.0  \\
AccAlign + AccAlign          & X + D   & WS- + \twstok & NLLB & 56.9 & 79.5 & 75.2 & 73.3 & 79.0 & 64.2 & 81.7 & 71.5 & 59.3 & 75.6 & 76.4 & 83.9 & 79.3 & 67.5 & 68.3 & 72.8 & 53.0 & 79.1 & 72.0   \\
AccAlign + AccAlign          & X + D   & WS- + \tsptok & NLLB & 57.0 & 79.2 & 76.4 & 73.5 & 80.6 & 64.7 & 82.7 & 71.9 & 60.2 & 76.0 & 77.5 & 84.2 & 79.9 & 68.3 & 68.8 & 72.9 & 53.9 & 79.1 & 72.6   \\
AccAlign + AccAlign          & X + D   & WS- + \tsptok & GT   & 61.0 & 79.0 & -    & 73.6 & 81.1 & 64.9 & -    & -    & -    & -    & 78.5 & 85.0 & -    & 67.8 & -    & 74.0 & 58.2 & 80.0 & -    \\
AccAlign + AccAlign          & X + ShL & WS- + \tsptok & NLLB & 55.6 & 76.5 & 73.4 & 73.0 & 75.7 & 60.5 & 78.9 & 66.4 & 58.6 & 71.3 & 73.1 & 83.3 & 78.1 & 64.3 & 66.5 & 70.5 & 46.3 & 76.3 & 69.3   \\
AccAlign + AccAlign          & X + L   & WS- + \tsptok & NLLB & 53.9 & 77.2 & 74.2 & 72.9 & 74.2 & 60.5 & 80.0 & 67.1 & 58.2 & 72.1 & 74.1 & 83.1 & 78.0 & 63.8 & 67.2 & 70.5 & 47.2 & 77.5 & 69.5   \\
Easy + AccAlign   & X + D   & - + \tsptok & NLLB & 58.0 & 79.2 & 74.0 & 73.3 & 76.4 & 64.5 & 83.2 & 71.1 & 56.3 & 77.8 & 72.6 & 85.2 & 78.7 & 68.6 & 68.3 & 73.8 & 53.6 & 76.3 & 71.7   \\
Codec + AccAlign  & X + D   & - + \tsptok & NLLB & 55.4 & 79.2 & 74.9 & 73.5 & 77.7 & 63.4 & 82.4 & 72.2 & 57.9 & 77.4 & 77.6 & 85.0 & 77.9 & 69.3 & 69.4 & 74.0 & 54.3 & 79.1 & 72.3   \\ \bottomrule
\end{tabular}
\caption{Main results for translation-based XLT evaluated on Masakha using different \was, pre-tokenizations, and MT models. We use XLM-R (X), DeBERTa (D), LLM2Vec Sheared-Llama 1.3B (ShL), and LLM2Vec LLama 3 8B (L). 
}
\end{table*}

\begin{table*}[t!]
\scriptsize
\centering
\setlength{\tabcolsep}{5.1pt}
\begin{tabular}{@{}llllcccccccccccc@{}}
\toprule
             &       &               &      & ar   & da   & de   & de-st & id   & it   & kk   & nl   & sr   & tr   & zh   & Avg  \\ \midrule
\zs           &       &               &      & 71.5 & 85.6 & 80.8 & 43.9  & 86.8 & 88.2 & 80.8 & 88.8 & 79.0 & 81.5 & 57.4 & 76.8 \\ \midrule
\multicolumn{16}{c}{\textit{\textbf{Translate-Train}}}                                                                                  \\ \midrule
AccAlign          & X     & \twstok        & NLLB & 82.6 & 76.0 & 86.1 & 62.2  & 87.4 & 88.1  & 86.0 & 85.5 & 85.0 & 85.3 & 85.1 & 82.7 \\
AccAlign$^\dag$     & X     & \twstok        & NLLB & 81.8 & 76.4 & 87.7 & 63.8  & 82.9 & 87.8  & 85.7 & 85.4 & 86.6 & 87.2 & 86.7 & 82.9 \\
Awesome$^\dag$ & X     & \twstok        & NLLB & 79.1 & 77.1 & 85.6 & 61.3  & 82.7 & 87.3  & 74.3 & 85.8 & 84.5 & 77.7 & 82.4 & 79.8 \\
Easy         & X     & -        & NLLB & 83.0 & 84.0 & 89.4 & 62.2  & 86.3 & 87.5  & 89.2 & 88.3 & 81.4 & 86.3 & 80.5 & 83.4 \\
Codec        & X     & -        & NLLB & 81.9 & 84.6 & 88.7 & 62.5  & 89.8 & 88.5  & 85.1 & 89.9 & 82.7 & 81.2 & 84.4 & 83.6 \\ \midrule
\multicolumn{16}{c}{\textit{\textbf{Translate-Test}}}                                                                                   \\ \midrule
AccAlign                        & X     & \tsptok        & NLLB & 78.8 & 76.0 & 86.3 & 60.9  & 78.8 & 88.1 & 82.5 & 87.5 & 79.8 & 81.3 & 82.3 & 80.2 \\
AccAlign$^\dag$                   & X     & \tsptok        & NLLB & 77.8 & 75.4 & 84.8 & 59.6  & 79.3 & 85.7 & 82.1 & 86.8 & 80.0 & 82.0 & 82.6 & 79.6 \\
Awesome$^\dag$              & X     & \tsptok        & NLLB & 73.8 & 74.9 & 84.8 & 59.2  & 71.0 & 84.8 & 63.0 & 87.2 & 77.1 & 70.8 & 76.4 & 74.8 \\
AccAlign                        & D     & \tsptok        & NLLB & 79.3 & 75.8 & 85.6 & 59.1  & 80.0 & 88.7 & 82.5 & 86.7 & 80.0 & 82.2 & 82.1 & 80.2 \\
AccAlign                        & X     & \twstok        & NLLB & 72.9 & 70.6 & 79.1 & 55.7  & 72.2 & 81.2 & 74.2 & 82.0 & 73.8 & 72.9 & 72.6 & 73.4 \\
AccAlign                        & D     & \twstok        & NLLB & 73.4 & 70.2 & 77.9 & 53.6  & 73.3 & 82.2 & 74.2 & 80.7 & 73.8 & 73.4 & 72.5 & 73.2 \\
AccAlign                             & X     & \tsptok        & GT   & 80.5 & 76.5 & 86.4 & 61.8  & 78.6 & 88.2 & 83.8 & 87.0 & 82.2 & 81.6 & 86.2 & 81.2 \\
AccAlign                        & D     & \tsptok        & GT   & 80.5 & 76.5 & 85.6 & 59.3  & 79.6 & 89.3 & 84.0 & 87.1 & 83.4 & 82.4 & 85.7 & 81.2 \\
AccAlign                        & ShL   & \tsptok        & NLLB & 68.4 & 68.3 & 76.8 & 51.6  & 68.6 & 77.7 & 72.2 & 78.1 & 70.7 & 72.2 & 73.3 & 70.7 \\
AccAlign                        & L     & \tsptok        & NLLB & 68.6 & 69.2 & 76.1 & 50.1  & 69.9 & 76.9 & 70.4 & 77.8 & 70.8 & 71.1 & 72.8 & 70.3 \\
Codec                      & X     & \tsptok        & NLLB & 79.0 & 81.9 & 86.1 & 60.4  & 84.8 & 88.4 & 83.0 & 86.5 & 72.4 & 83.6 & 67.0 & 79.4 \\
Codec                      & D     & \tsptok        & NLLB & 79.9 & 81.8 & 85.5 & 58.8  & 85.8 & 89.0 & 83.2 & 86.0 & 72.9 & 84.2 & 67.5 & 79.5 \\ \midrule
\multicolumn{16}{c}{\textit{\textbf{Ensembling-Translate-Train}}}                                                                              \\ \midrule
AccAlign + AccAlign                        & X + X   & WS- + \twstok & NLLB & 80.9 & 75.1 & 85.9 & 63.1  & 88.1 & 89.2 & 81.4 & 86.3 & 82.7 & 81.5 & 80.4 & 81.3 \\
AccAlign + AccAlign                        & X + X   & WS- + \tsptok & NLLB & 82.0 & 76.3 & 88.2 & 64.5  & 88.7 & 89.6 & 86.5 & 86.4 & 82.9 & 85.0 & 85.0 & 83.2 \\
AccAlign + AccAlign                        & X + D   & WS- + \twstok & NLLB & 81.2 & 75.0 & 85.8 & 61.7  & 88.2 & 89.9 & 81.1 & 86.6 & 82.6 & 81.7 & 79.6 & 81.2 \\
AccAlign + AccAlign                        & X + D   & WS- + \tsptok & NLLB & 82.4 & 76.1 & 88.2 & 64.2  & 89.0 & 90.0 & 86.6 & 87.2 & 83.1 & 85.3 & 84.8 & 83.4 \\
AccAlign + AccAlign                        & X + D   & WS- + \tsptok & GT   & 82.4 & 75.8 & 87.1 & 64.3  & 87.7 & 89.8 & 86.5 & 86.3 & 84.4 & 84.2 & 86.4 & 83.2 \\
AccAlign + AccAlign                        & X + ShL & WS- + \tsptok & NLLB & 81.2 & 76.2 & 87.4 & 63.6  & 87.7 & 89.6 & 86.0 & 86.1 & 83.7 & 84.7 & 84.5 & 82.8 \\
AccAlign + AccAlign                        & X + L   & WS- + \tsptok & NLLB & 79.7 & 76.4 & 87.3 & 61.8  & 86.1 & 88.4 & 85.4 & 86.1 & 82.9 & 83.3 & 84.0 & 81.9 \\
Easy + AccAlign                 & X + D   & - + \tsptok & NLLB & 81.1 & 81.2 & 89.7 & 64.4  & 89.0 & 90.6 & 85.7 & 90.4 & 82.7 & 82.4 & 84.5 & 83.8 \\
Codec + AccAlign                & X + D   & - + \tsptok & NLLB & 82.5 & 75.3 & 89.1 & 62.7  & 87.6 & 88.7 & 87.1 & 89.2 & 81.8 & 85.3 & 81.3 & 82.8 \\ \bottomrule
\end{tabular}
\caption{Main results for translation-based XLT evaluated on xSID using different \was, pre-tokenizations, and MT models. We use XLM-R (X), DeBERTa (D), LLM2Vec Sheared-Llama 1.3B (ShL), and LLM2Vec LLama 3 8B (L). 
}
\end{table*}

\begin{table*}[t!]
\section{Further Results: Comparison of Projection Methods}
\label{app:ablation_proj}
\small
\centering
\setlength{\tabcolsep}{13.7pt}
\begin{tabular}{@{}llccccccccc@{}}
\toprule
 &  & hau  & ibo  & nya  & sna  & swa  & xho  & yor  & zul  & Avg  \\ \midrule
\ept & X & 73.0 & 64.6 & 68.3 & 57.2 & 84.1 & 71.3 & 37.0 & 70.6 & 65.8 \\
\ept & mD & 73.1 & 66.9 & 74.5 & 48.9 & 84.4 & 69.3 & 35.6 & 70.6 & 65.4 \\
CLaP & X & 68.2 & 50.5 & 63.5 & 66.5 & 77.0 & 63.1 & 33.6 & 63.4 & 60.7 \\
CLaP & mD & 65.4 & 51.1 & 72.8 & 72.0 & 77.0 & 63.0 & 33.5 & 62.7 & 62.2 \\
T-Proj & X & 71.9 & 64.2 & 66.0 & 68.7 & 83.7 & 68.4 & 37.8 & 70.5 & 66.4 \\
T-Proj & mD & 72.2 & 67.7 & 75.7 & 73.7 & 82.9 & 68.8 & 40.6 & 70.2 & 69.0 \\
\codect & X & 73.4 & 65.5 & 68.3 & 70.9 & 84.2 & 70.2 & 39.4 & 75.3 & 68.4 \\
\codect & mD & 72.5 & 67.1 & 74.9 & 73.2 & 82.8 & 69.2 & 41.1 & 72.4 & \textbf{69.2} \\
\wa & X & 73.1 & 72.2 & 66.2 & 70.9 & 83.2 & 69.6 & 40.2 & 75.6 & 68.9 \\
\wa & mD & 72.7 & 68.7 & 75.6 & 73.0 & 81.8 & 69.4 & 40.6 & 71.7 & \textbf{69.2} \\ \bottomrule
\end{tabular}
\caption{Results for \ttr on Masakha using different projection methods and downstream models---XLM-R (X) and mDeBERTa (mD). For \wa, we use the same setting as for our main results. We subset the languages to those seen in the pretraining of mT5 which is the backbone model of T-Proj \cite{garcia-ferrero-etal-2023-projection}.
}
\end{table*}

\end{document}